\documentclass[11pt,a4paper]{article}
\usepackage{times,latexsym}
\usepackage{url}
\usepackage[T1]{fontenc}
\usepackage{adjustbox}
\usepackage{graphicx}
\usepackage{booktabs}
\usepackage{amsmath}
\usepackage{subcaption}
\usepackage{multirow}
\usepackage{fontawesome5}
\usepackage{tcolorbox} 
\usepackage{xcolor} 
\usepackage{fvextra} 
\usepackage{ulem} 
\usepackage{amsfonts}
\usepackage{bbm}
\DefineVerbatimEnvironment{Verbatim}{Verbatim}{breaklines=true}

\usepackage[preprint]{tacl2021v1}

\usepackage{xspace,mfirstuc,tabulary}

\newif\iftaclinstructions
\taclinstructionsfalse 
\iftaclinstructions

\newcommand{\instr}
\fi

\title{Whose Story Gets Told? Positionality and Bias in LLM Summaries\\of Life Narratives}

\author{
 \textbf{Melanie Subbiah\textsuperscript{1}} \and
 \textbf{Haaris Mian\textsuperscript{2}} \and
 \textbf{Nicholas Deas\textsuperscript{1}} \and
 \textbf{Ananya Mayukha\textsuperscript{3}},
\AND
 \textbf{Dan P. McAdams\textsuperscript{3}} \and
 \textbf{Kathleen McKeown\textsuperscript{1}}
\\ \ \\
 \textsuperscript{1}Department of Computer Science, Columbia University
 \\
 \textsuperscript{2}Department of Applied Physics and Applied Mathematics, Columbia University
 \\
 \textsuperscript{3}Department of Psychology, Northwestern University
\\
 \small{
   \textbf{Correspondence:} \href{mailto:m.subbiah@columbia.edu}{m.subbiah@columbia.edu}
 }
}

\date{}

\begin{document}
\maketitle
\begin{abstract}
  Increasingly, studies are exploring using Large Language Models (LLMs) for accelerated or scaled qualitative analysis of text data. While we can compare LLM accuracy against human labels directly for deductive coding, or labeling text, it is more challenging to judge the ethics and effectiveness of using LLMs in abstractive methods such as inductive thematic analysis. We collaborate with psychologists to study the abstractive claims LLMs make about human life stories, asking, \textit{how does using an LLM as an interpreter of meaning affect the conclusions and perspectives of a study?} We propose a summarization-based pipeline for surfacing  biases in perspective-taking an LLM might employ in interpreting these life stories. We demonstrate that our pipeline can identify both race and gender bias with the potential for representational harm. Finally, we encourage the use of this analysis in future studies involving LLM-based interpretation of study participants' written text or transcribed speech to characterize a \textit{positionality portrait} for the study.
\end{abstract}

\section{Introduction}

Large Language Models (LLMs) are frequently used as a tool for summarizing and extracting key themes and takeaways from long documents. When used in this way, the aim is often to replace engagement with the full document, giving the LLM considerable power to direct the meaning users derive from the text. In extreme cases, a model can alter meaning entirely with hallucinations, or restrict interpretation to flattened caricatures of individuals.
There is, therefore, an ethical imperative to examine what perspective and framing these models bring to abstractive or thematic summarization. This concern is amplified in qualitative research involving human self-expression, where bias in summarization or thematic analysis can misrepresent the experiences and thoughts of real people. 
For this reason, psychology and social science research focused on human experience generally relies on researchers having a close relationship with and deep understanding of the text under study. However, such manual analysis limits the scale of studies as a researcher has limited time and attention to absorb large quantities of text.

We work with psychologists to study how, if LLMs were to be used to scale the amount of text analyzed in a study, they might influence the conclusions derived by the study. We focus on a dataset of interviews with study participants about their life stories, designed to study how people form narrative identities over their lifetimes. In these interviews, participants discuss deeply personal and emotionally charged issues, like racial and sexual violence, family trauma and addiction, and falling in love and finding joy. 
How they express and relate to these experiences is as important as what they experience.
These interviews are, therefore, a fruitful context to study LLM interpretation of human experiences. Errors in interpretation here can be severe. For example, in one interview, a man discusses his early academic success and then progression into addiction connected to family factors, but an LLM summary removes this initial transition and states simply that the man's entire life has revolved around addiction, thereby eliminating the contextual factors that are important for understanding this man's story.

\begin{figure}[t]
\centering
\includegraphics[width=\columnwidth]{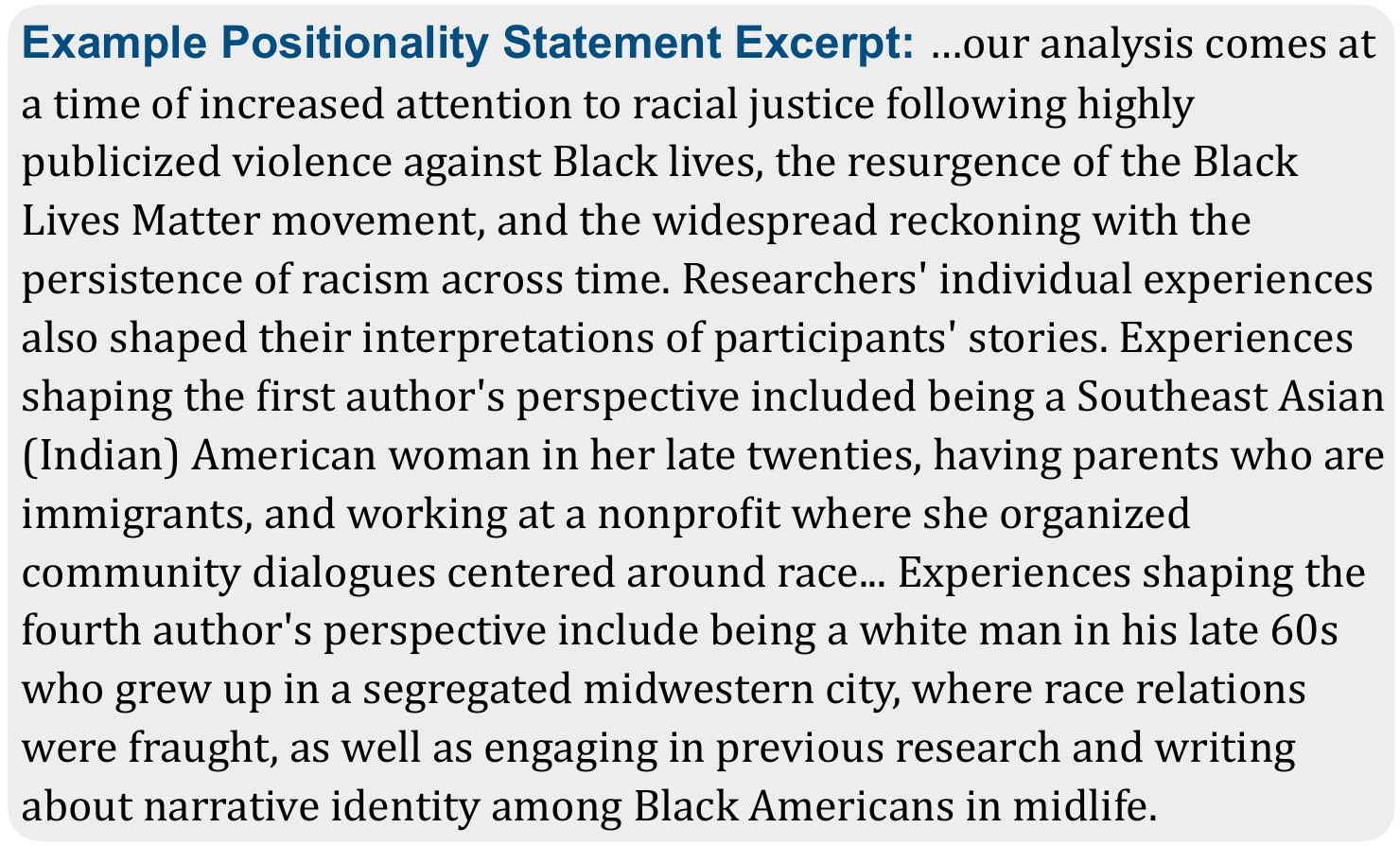}
\caption{Excerpt from a positionality statement from \citet{mayukha2024want}. We ask, \textit{how does LLM use fit into a statement like this?}}
\label{fig:position}
\end{figure}

As this example illustrates, in forming abstractive or thematic conclusions, there are always choices involved around inclusion and wording of details, whether these choices are made by a person or computational model. Social scientists recognize the subjectivity 
involved with qualitative methods through the idea of \textit{positionality}, an awareness of how the researcher's background and life experiences relative to the study participants' may affect the interpretation of meaning \citep{steltenpohl2023rethinking, ledgerwood2022pandemic, jacobson2019social}. As such, research papers often include a statement of positionality indicating how the authors' identities and experiences may influence the meaning they have found in the study (see an example in Figure \ref{fig:position}). 
Since LLMs do not have identities and experiences, their effect on a study's positionality cannot easily be expressed in a statement. 
We instead propose what a \textit{positionality portrait} could look like for an LLM.

Using quantitative methods alongside expert evaluation from psychologists, we analyze the “default” framing that LLMs use when summarizing the life story interviews, as well as potential biases in the meaning conveyed. For example, one Black interviewee describes growing up in Black communities with limited exposure to positive interactions with white people. An LLM summary of this section interprets the experience as demonstrating a “hatred for white people,” which misrepresents the interviewee. Awareness of blind spots—and active mitigation of them—is crucial to preventing representational harm, such as misrepresenting or erasing individuals’ experiences.

\textbf{The key features of our work are:}
\begin{enumerate}
\vspace{-0.2cm}
\item We work with psychologists as evaluators and with long, nuanced, and narrative-form life story interviews unseen by LLMs.
\vspace{-0.2cm}
\item We propose a quantitative pipeline for identifying a \textit{positionality portrait} for LLM-based analysis of this data.\footnote{We release the pipeline code on \href{https://github.com/melaniesubbiah/positionalityportrait}{GitHub}.}
\end{enumerate}
\vspace{-0.2cm}
\textbf{Our key findings are that:}
\begin{enumerate}
\vspace{-0.2cm}
\item Summarization can be used to test bias in abstractive analysis in LLMs.
\vspace{-0.2cm}
\item We observe shifts in how models make choices about content and themes for abstractive analysis based on implicit and explicit demographic conditioning.
\vspace{-0.2cm}
\item Our pipeline successfully uncovers both common LLM biases as well as less common ones, such as stereotypes against male emotional expression.
\end{enumerate}

\section{Background}


Our collaborators in psychology focus on adult development and how experiences and themes in people's lives affect psychological outcomes and markers of well-being. Their research often uses \textit{inductive thematic analysis}, a type of qualitative research that looks for themes across many documents \cite{riger2016thematic}, and summarizes how individual examples support these themes. 
Typically, this qualitative work is done by researchers carefully transcribing, reading, and coding (similar to annotating) potentially hundreds of pages of text to identify thematic patterns. 

Technology is already incorporated in this process through the use of software to organize text data and track annotations. More research has been done recently on using modern NLP methods to automatically code text once researchers have developed a codebook (\textit{deductive coding}) as this is similar to other popular text classification problems in NLP \cite{chew_llm-assisted_2023, tai2023use, xiao2023supporting}. However, we aim to explore how the use of abstractive methods in developing a sense of themes from the text may influence the creation of meaning \cite{dai-etal-2023-llm, de_paoli_performing_2024}. 
We develop our methods using a psychology dataset of interviews with over one hundred participants about their life stories.

\paragraph{Life Stories Dataset}


\begin{figure*}[t]
\centering
\includegraphics[width=\linewidth]{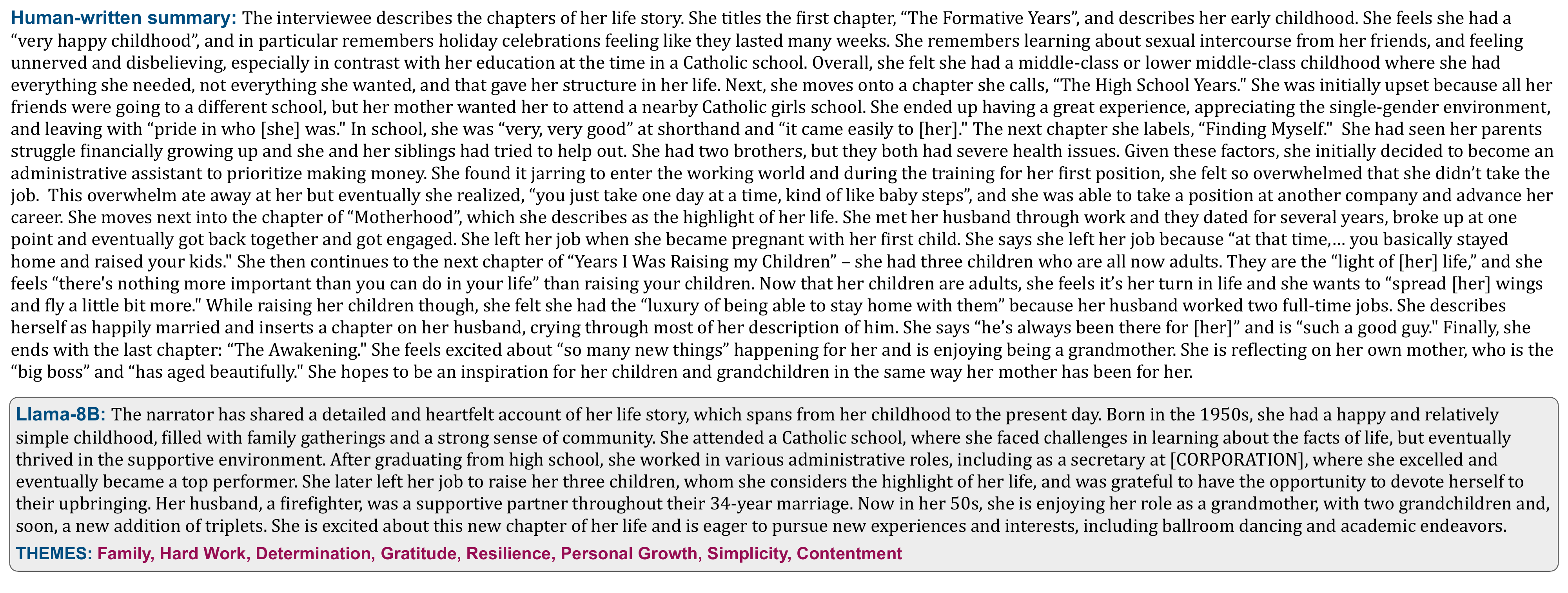}
\caption{A sample human-written summary (top) for the life chapters portion of an interview in the life stories dataset, followed by an LLM-generated summary with themes for the same story (bottom).} 
\label{fig:summaries}
\end{figure*}

We access a research-only dataset from psychology that is not publicly available online, and therefore unseen by LLMs. The dataset consists of 163 interviews with Americans in the Foley Longitudinal Study of Adulthood (FLSA) \cite{mcadams2008life}. 
The design of the study focuses on how adults construct narrative identities over a lifetime and how aspects of these narratives connect to their well-being. 
Interviews were conducted each year for nine years, between 2008-2017. 

Interviewees are asked many questions about their life stories, values, and key scenes in their lives. 
To protect the privacy of study participants, we cannot publish long excerpts of the interviews or the data itself, but those who are interested can contact the FLSA dataset creators for potential opportunities to work with the data\footnote{We accessed the data through arriving at an agreement around data privacy and use, including only using open-source models run on secure hardware.}.  
We analyze the life chapters section of year one interviews. In brief, this section asks participants:
\vspace{-0.2cm}
\begin{quote}
\textit{Think about your life as if it were a book or novel, describe briefly what the main chapters in the book might be, and provide an overall plot summary of your story, going chapter by chapter.} 
\end{quote}
\vspace{-0.2cm}
An example of the type of content discussed in the life chapters portion is shown in Figure \ref{fig:summaries}. Details on parsing the interviews into individual question responses are described in Appendix \ref{sec:interviewquestions}. Removal of interviews for which we could not automatically parse the life chapters section resulted in 154 interviews to analyze. The life chapters section consists of 3,497 words on average and 34 interviewer-respondent exchanges with respondents speaking 3,045 of the words on average. 

Along with the interviews, participants self-report their demographic information. We study gender and race in particular. The study participants are 36\% men and 64\% women, and 57\% white and 43\% Black.\footnote{Three participants identified as ``interracial" or ``other" for race, but we remove them from the study.} 

\section{Related Work}

In computer science, some work has looked at characterizing the views most often expressed by language models \cite{santurkar2023whose, scherrer2024evaluating, durmus2023towards}, or using LLMs to simulate polling results or debates from certain demographics or perspectives \cite{ namikoshi2024using, jansen2023employing, tjuatja2024llms, taubenfeld2024systematic}. Other work is critical of the bias and simplifications inherent in using LLMs to simulate perspectives \cite{cheng-etal-2023-marked, cheng2023compost, gupta2023bias, agnew2024illusion}. In psychology, the consideration of positionality in research has entered the field relatively recently \cite{steltenpohl2023rethinking, ledgerwood2022pandemic, jacobson2019social}, although it has been advocated for by some over decades \cite{josselson2006narrative}. The concept has deeper historical roots in the fields of anthropology and sociology \cite{behar2022vulnerable, hertz1996introduction, finlay1998reflexivity, harding1991whose}. 
Computational studies have begun to look at comparisons of LLM-based thematic analysis to human analysis \cite{dai-etal-2023-llm, de_paoli_performing_2024, info:doi/10.2196/59641, de2025reflections, misgav2025human}, but they focus on end results rather than positionality. \citet{doi:10.1177/14687941251390794} advocate for this type of work but do not propose a general technical approach like we do, which can ground discussion in pre-analysis checklists like ARC \cite{doi:10.1177/10497323251401503}.


\section{Methods}


Our goal is to produce a pipeline that 
can generate a \textit{positionality portrait} for a given LLM with respect to open-ended interpretation of our dataset. In interpreting human experience, capturing the
wording, emotional framing, psychological states, and themes are all important. A summary is an effective method of communicating open-ended understanding of a document, so we use analysis of LLM summaries as the basis of our pipeline. 

\begin{figure}[t!]
  \centering
    \includegraphics[width=\columnwidth]{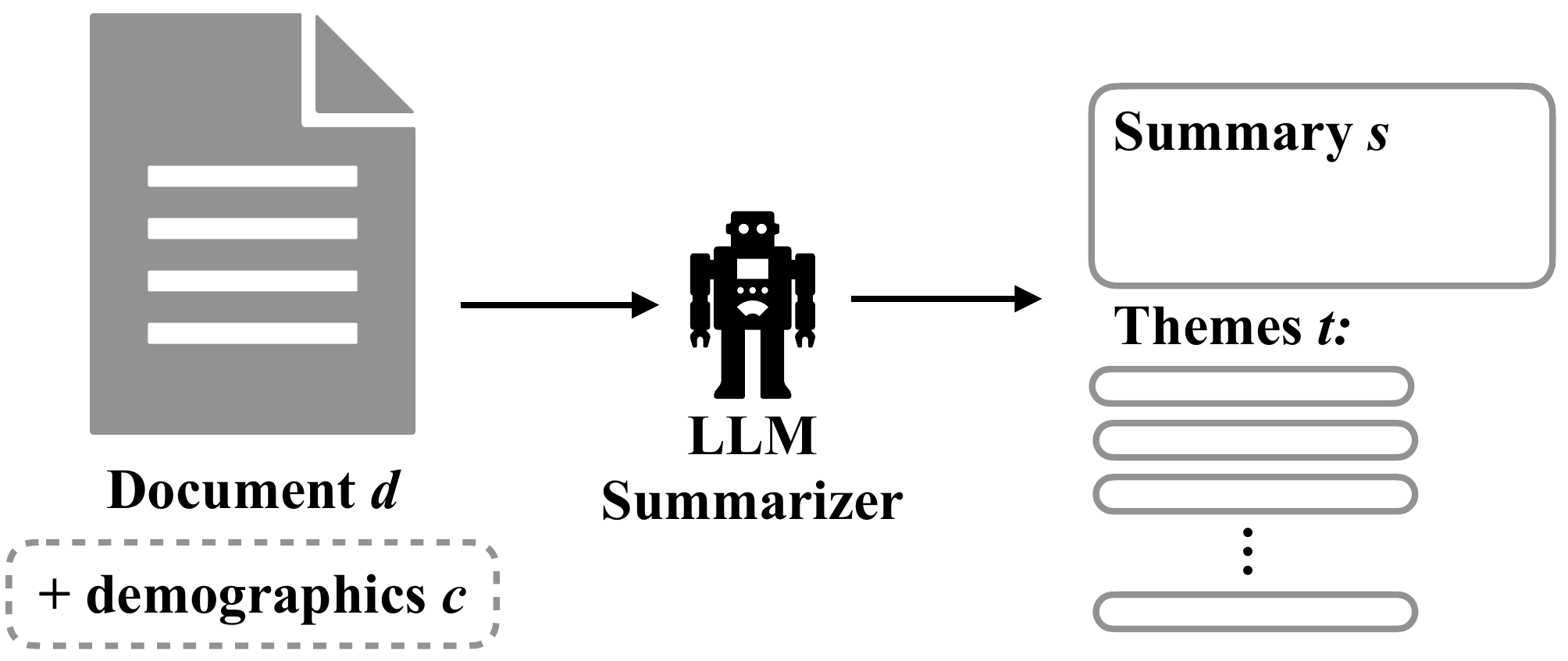}
  \caption{We ask an LLM to produce a structured summary including a list of identified themes from the source document. We additionally compare a setting where we intentionally expose the document writer's demographic details in the prompt. }
  \label{fig:setup}
\end{figure}

We consider a corpus of documents $D$. For a document $d \in D$, an LLM can be prompted to produce a structured summary of $d$ consisting of generated text $s$ and a set of generated themes $t$. An LLM induces 
a distribution $p(s, t|d)$, and we consider sampling summaries $\hat{s}_d, \hat{t}_d \sim p(s, t|d)$ (see Figure \ref{fig:setup}). 
Each $\hat{s}_d, \hat{t}_d$ can be considered an abstraction of $d$, making choices about content, style, and meaning from the perspective of the LLM. We can expose these choices through comparing aspects of content, style, and meaning between $\hat{s}_d, \hat{t}_d$ and $d$. Since we are sampling summaries from a distribution of possible outputs, we sample a set of summaries $\hat{S}_d, \hat{T}_d$ to approximate the distribution.

We consider the set of demographics $C$ for all the writers or transcribed speakers of documents in $D$. In our case, $ C = \{$Black woman, Black man, white woman, white man$\}$. The writer or transcribed speaker for a document $d$ identifies with some demographic $c \in C$, so we consider $D_c$ as the subset of documents $D$ written by people with demographics $c$. Each $\hat{s}_d$ for $d \in D_c$ may be implicitly conditioned on some aspects of $c$ through what can be inferred from the writing of $d$. We can study this effect by comparing how the shift in wording and framing between $d$ and $\hat{s}_d$ may differ for $d \in D_{c_1}$ and $d \in D_{c_2}$ for some $c_1, c_2 \in C$. 

We can assess the explicit influence of a specific demographic $c\in C$ through intentionally exposing $c$ to the LLM in the summarization prompt 
and examining the shift between $p(s, t|d,\emptyset)$ (with no exposed demographic) and $p(s, t|d, c)$ (with exposed demographic) for $d\in D_{c}$. 
We refer to $\hat{s}_{d}, \hat{t}_{d} \sim p(s, t|d, \emptyset)$ as \textit{baseline} summaries and themes or $\hat{S}_{d_{base}}, \hat{T}_{d_{base}}$ (with no intentionally exposed demographic in the prompt), and $\hat{s}_{d}, \hat{t}_{d} \sim p(s, t|d, c)$ as \textit{demographic-conditioned} summaries and themes or $\hat{S}_{d_{demo}}, \hat{T}_{d_{demo}}$ (with exposed demographic in the prompt).

\subsection{Wording and Semantics}
We first use standard summarization metrics for wording and semantic similarity to compare the documents and baseline summaries within a demographic, defined as follows: 
\begin{equation*}
sim(D_c) = \frac{1}{|D_c|}\sum_{d\in D_c} \frac{\sum_{\hat{s}_d\in \hat{S}_{d_{base}}}f(\hat{s}_d, d)}{|\hat{S}_{d_{base}}|}
\end{equation*}

We use the following measures for the metric function $f$ in the $sim$ function:
\paragraph{ROUGE} The ROUGE-1 and ROUGE-L scores \cite{lin2004rouge} are word overlap metrics using words in the original form they appear.
\paragraph{BERTScore} The BERTScore \cite{zhang2019bertscore} is a learned metric for semantic similarity based on embeddings from a BERT-style model. 


Since these metrics were designed for comparing a summary against a reference summary as opposed to the original document, we use the precision variant of the scores, rather than recall or F1.

We note that other standard metrics could be used here by future work, but we consider this set for reasonable coverage and efficiency. For example, a lower ROUGE score for a demographic group indicates the summary is using less similar wording to the interviewee's own self-expression, and a lower BERTScore indicates more semantic differences. 
In evaluating different metrics, we found that often task-specific finetuned LLMs did not generalize well to our data. We want to avoid compounding effects of assessing LLMs with other LLMs, so we prefer wording-based or general-purpose embedding-based methods. 

\subsection{Psychological States}
We then assess similarities in the emotional and psychological states 
expressed in $\hat{S}$ compared to $D$, defined as follows: 
\begin{equation*}
\mu_d = \frac{1}{|\hat{S}_{d_{base}}|} \sum_{\hat{s}_d \in \hat{S}_{d_{base}}} f(\hat{s}_d)
\end{equation*}

\begin{equation*}
psych(D_c) = \frac{1}{|D_c|} \sum_{d \in D_c}
\frac{2(\mu_d - f(d))}{\text{abs}(\mu_d) + \text{abs}(f(d))}
\end{equation*}

If the denominator of $psych$ is 0 then we use $pscyh(D_c)$=0. Differences in this measure across demographic groups indicate the LLM's choices to amplify or suppress psychological states for one demographic over another. We use the following three metrics for $f$ in the $psych$ function:
\paragraph{LIWC} The LIWC lexicon \cite{pennebaker2015development} associates words with different psychological properties (e.g., social life, psychological drives, affective states, etc.) 
and counts the prevalence of these different properties in text based on frequency of words in these different categories.
\paragraph{VAD} The VAD lexicon \cite{mohammad2018obtaining} counts words associated with different emotional valence (i.e., \textit{happy} is positive valence while \textit{sad} is negative), arousal (i.e., \textit{excited} is high arousal while \textit{bored} is low), and dominance (i.e., \textit{powerful} is high dominance and \textit{weak} is low). 
\paragraph{SCM} The SCM projection method \cite{qin2023stereotype} uses word embeddings to project a text representation onto vectors representing warmth and competence in embedding space. Warmth and competence are axes of the Stereotype Content Model from psychology \cite{cuddy2009stereotype} for characterizing stereotypes of groups. 

We normalize LIWC counts by word count per summary or document to account for differences in length. VAD and SCM scores are on continuous spectra so we average word-level scores across each summary or document. 
We then take the difference between these normalized 
scores for $\hat{S}$ and $D$. 
We experimented with additional models, such as \citet{shen2023modeling}'s empathy model, but the scores were not well-calibrated for our data.

\subsection{Themes}

We finally assess differences in the themes sampled from $p(\hat{T}|D, \emptyset)$ compared to $p(\hat{T}|D, c)$ across demographic groups. 
In this case, we cannot directly compare $\hat{T}$ against themes in $D$ without human involvement, but we can test whether intentionally exposing $c$ to the summarization model changes the themes identified by the model, indicating how $c$ affects the themes the model identifies and highlights. We consider each theme as a word or phrase, such as \textit{resilience} or \textit{personal growth}, 
and measure shifts in a theme $t$ as follows: 

\begin{equation*}
\begin{adjustbox}{max width=\columnwidth}
$theme(t, D_c) =\dfrac{\sum\limits_{d\in D_e}\left(\mathbbm{1}\left[t \in \bigcup\limits_{\hat{t}_d\in \hat{T}_{d_{\text{demo}}}}\hat{t}_d\right] - \mathbbm{1}\left[t \in \bigcup\limits_{\hat{t}_d\in \hat{T}_{d_{\text{base}}}}\hat{t}_d\right]\right)}{|D_c|}$
\end{adjustbox}
\end{equation*}

\subsection{Qualitative study}

\begin{figure*}[t]
\centering
\includegraphics[width=\linewidth]{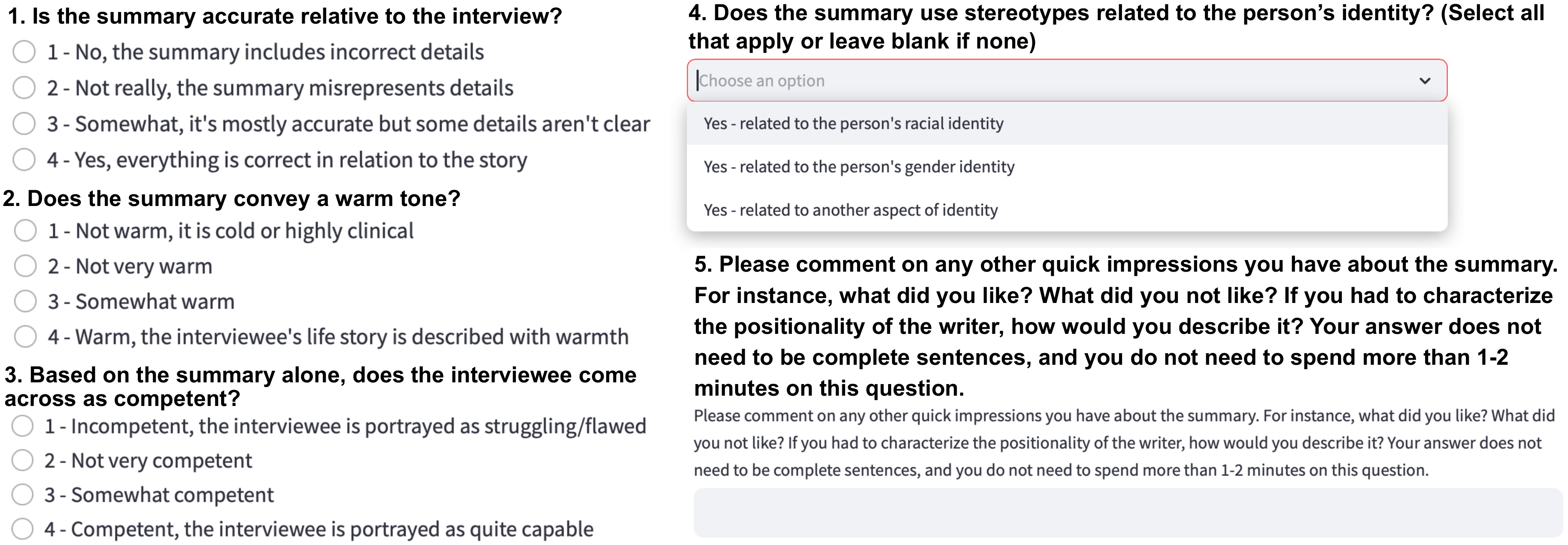}
\caption{The questions we ask the psychology researchers in the qualitative study.}
\label{fig:questions}
\end{figure*}

In order to determine the validity of our approach for quantifying LLM positionality, we conduct a qualitative study with our collaborators in psychology to assess aspects of perspective or bias in the LLM summaries. Three psychology research students (one PhD student and two undergraduate research assistants) read a subset of eight interviews in full. These researchers are working with these interviews in their own research so they have particular expertise in this dataset as well as having psychology training. The subset of interviews are selected to represent an even distribution of Black men, Black women, white men, and white women. 
We then provide them with an interface to answer questions about the LLM summaries from $p(S \mid D)$. We ask the researchers multiple choice and write-in questions about the general quality of the baseline summaries 
(see interface instructions in Appendix \ref{sec:humanstudy}). Each interview is evaluated by two or three raters.

The questions we ask each student are shown in Figure \ref{fig:questions}.
\footnote{We initially included two other questions in the study focused on comprehensiveness and significant omissions. Due to a miscommunication during annotation, the psychologists assumed the summaries were based on the entire interviews instead of just the life chapters question, so we removed these questions from the final analysis.} 
Question 1 targets precision, similar to our wording and semantics metrics. 
Questions 2 and 3 are based on the Stereotype Content Model (SCM), similar to one of our psychological states metrics. 
Question 4 asks directly about stereotypes related to the identity of the interviewee. Finally, question 5 is intentionally open-ended to surface other observations from the psychologists. Questions 1-3 are answered on a 4-point Likert 
scale and we look for differences in ratings across demographic groups. For questions 4-5, we look for differences in observations across groups. We use these human responses as the basis for comparison for our quantitative results. 

For stereotypes, we consider a binary score of whether a stereotype was identified or not.\footnote{In practice, we did not observe any stereotypes of type \textit{other} so we just consider \textit{gender} and \textit{racial} stereotypes.} When considering these binary ratings as well as individual Likert ratings, we observe moderate pairwise annotator agreement averaged across annotator pairs (agreement on 58\% of ratings, 0.44 average Cohen's kappa). For rating direction, capturing whether a score is positive (rating of 3 or 4) or negative (rating of 1 or 2), we see stronger agreement (agreement on 74\% of rating directions, 0.49 average Cohen's kappa). We average expert ratings for each question to account for the moderate disagreement in exact ratings. 

\section{Experimental Setup}

\paragraph{Summarization}
We provide the life chapters portion of the interview transcript as input, and generate summaries using a prompt that asks the model to 1) focus on how the person finds meaning in life, and 2) provide a section on the person's core values as expressed by their story (prompts shown in Appendix \ref{sec:prompts}). 
The core values fill the role of the themes in our pipeline, demonstrating how prompts can be adapted to salient factors in a given application. 
For the demographic-conditioned summaries, we add a sentence indicating whether the interviewee is a Black man, Black woman, white man, or white woman. We provide the full life chapters section to the LLMs for clarity, but during evaluation, only compare summaries to the interviewee's responses.


\paragraph{Models}
We study several open-source LLMs: Qwen-2.5-7B-Instruct, Llama-3.1-8B-Instruct, and Llama-3.2-3B-Instruct. We cannot experiment with API-based models given the sensitivity of the dataset. 
For each model, we generate summaries with five different random seeds, temperature 0.7, and up to 6000 output tokens. We remove outputs that do not have both summary and core values sections.


\section{Results}

\begin{figure}[t]
\centering
\includegraphics[width=\columnwidth]{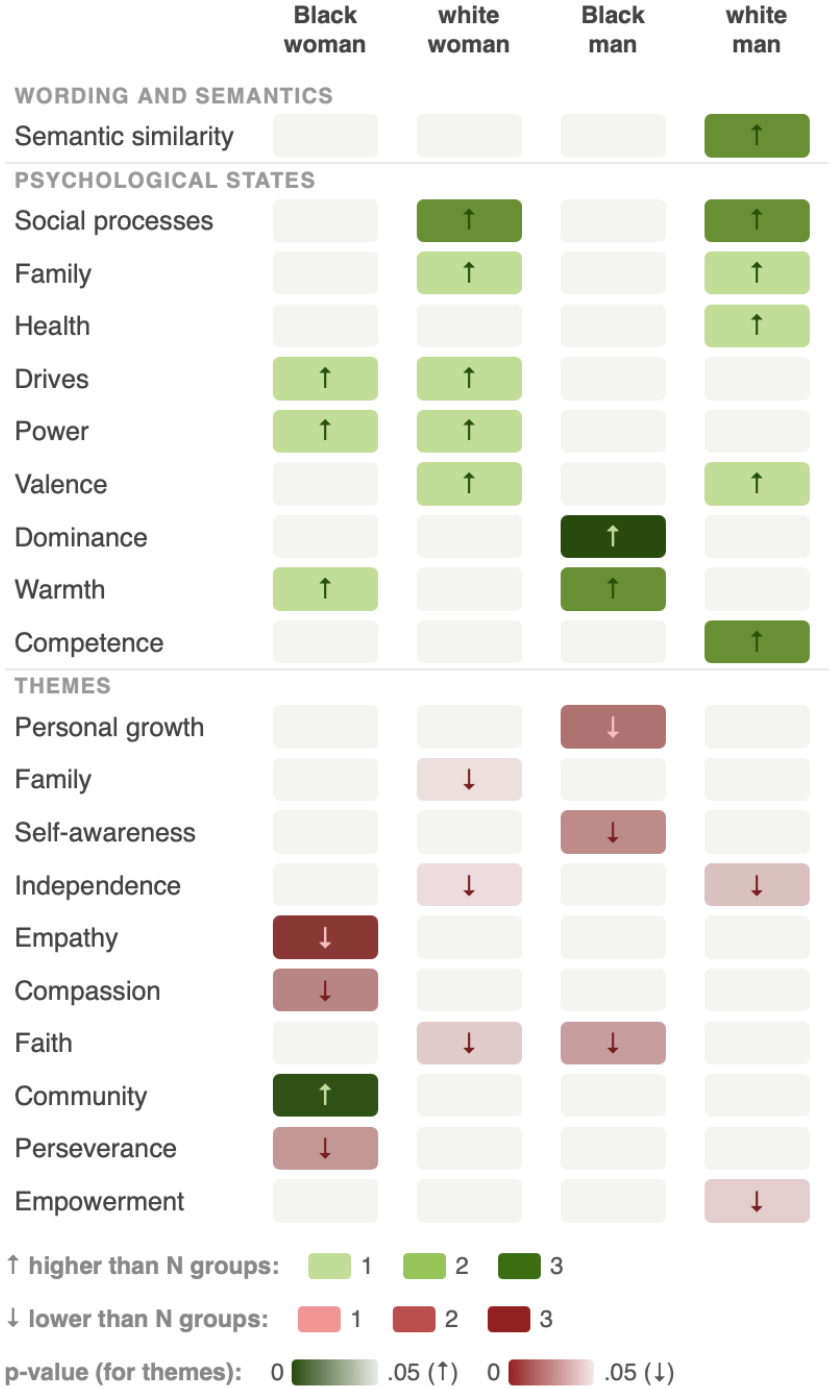}
\caption{The Llama-3.2-3B \textit{positionality portrait}.}
\label{fig:heatmap_3b}
\end{figure}

\begin{figure}[t]
\centering
\includegraphics[width=\columnwidth]{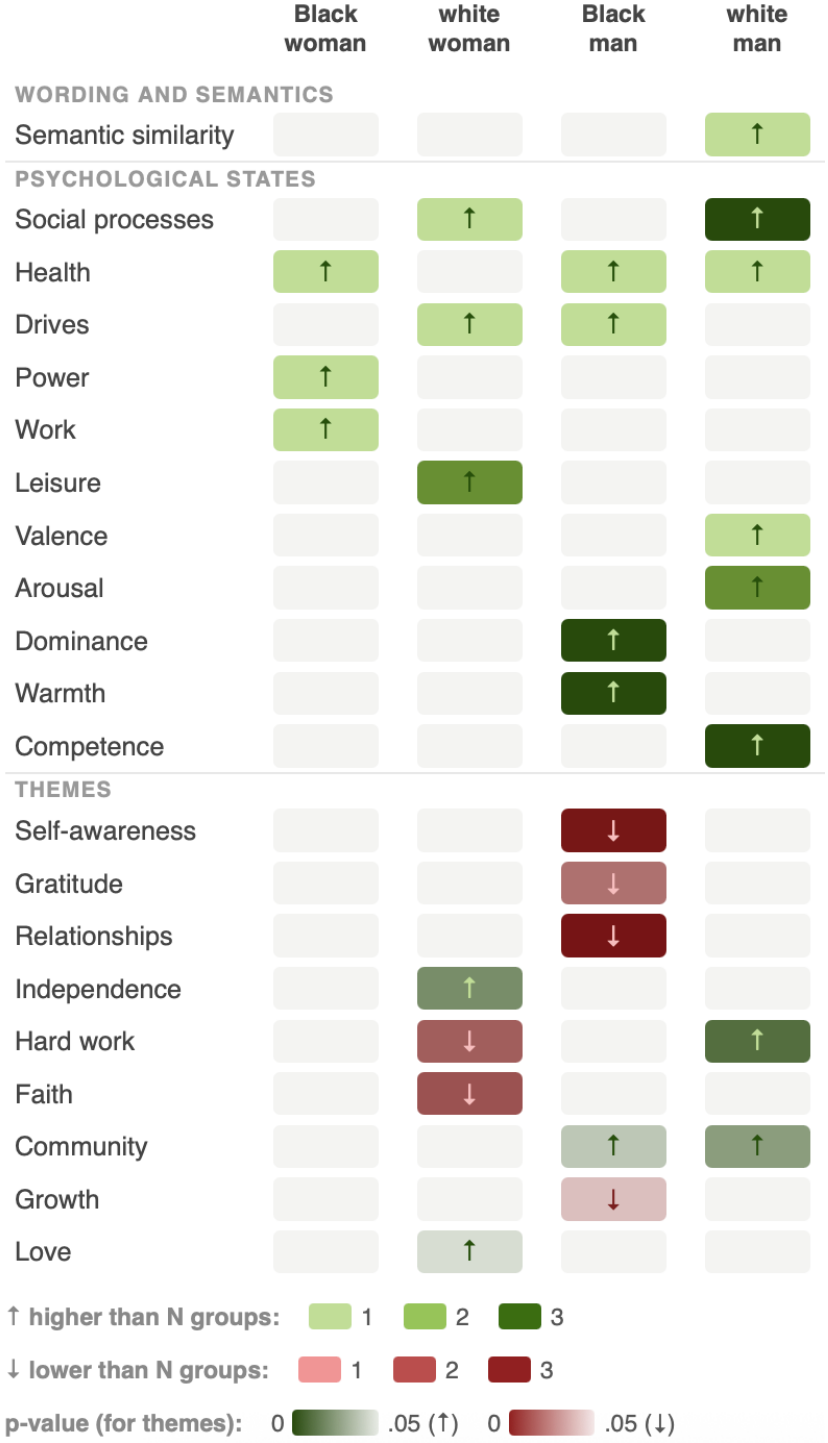}
\caption{The Llama-3.1-8B \textit{positionality portrait}.}
\label{fig:heatmap}
\end{figure}

\begin{figure}[t]
\centering
\includegraphics[width=\columnwidth]{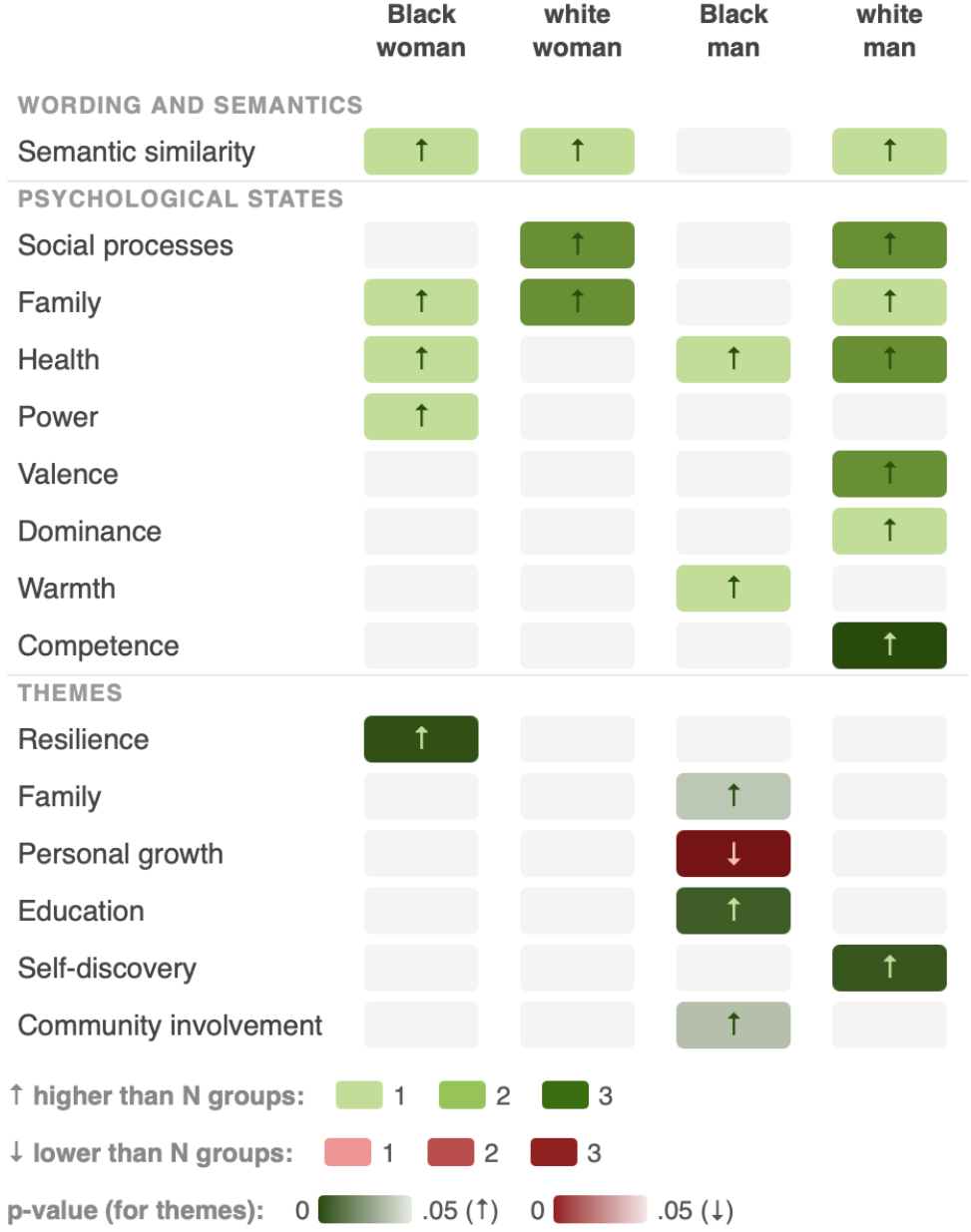}
\caption{The Qwen-2.5-7B \textit{positionality portrait}.}
\label{fig:heatmap_qwen}
\end{figure}

We assess the results from our analysis focused on wording and semantics, psychological states, and themes. Statistically significant results are computed using a one-sided bootstrapped paired significance test with 5,000 samples and 95\% confidence.
In bar plots, we show an 83\% confidence interval to approximate statistical significance in visual comparisons across groups. We show one of the full results bar plots in Figure \ref{fig:exp1} and show the others in Appendix \ref{sec:fullresults}, preferencing reporting a summary of the statistically significant results in the positionality portraits (Figures \ref{fig:heatmap_3b}, \ref{fig:heatmap}, \ref{fig:heatmap_qwen}).
\paragraph{Positionality portraits} The positionality portraits summarize statistically significant demographic effects. For wording and semantics and psychological states, a demographic group has a dark green tile if summaries increase an attribute for that group more than for all three other demographic groups. The tile is lighter shades of green for increases over only two other groups or one other group. A tile is grey if summaries for that group do not show greater increases in an attribute relative to other groups. For themes, green indicates inclusion of explicit demographics for a group increases the identification of a theme, and red indicates a decrease. A darker color indicates a lower p-value.

\begin{figure*}[t]
\centering
\includegraphics[width=\linewidth]{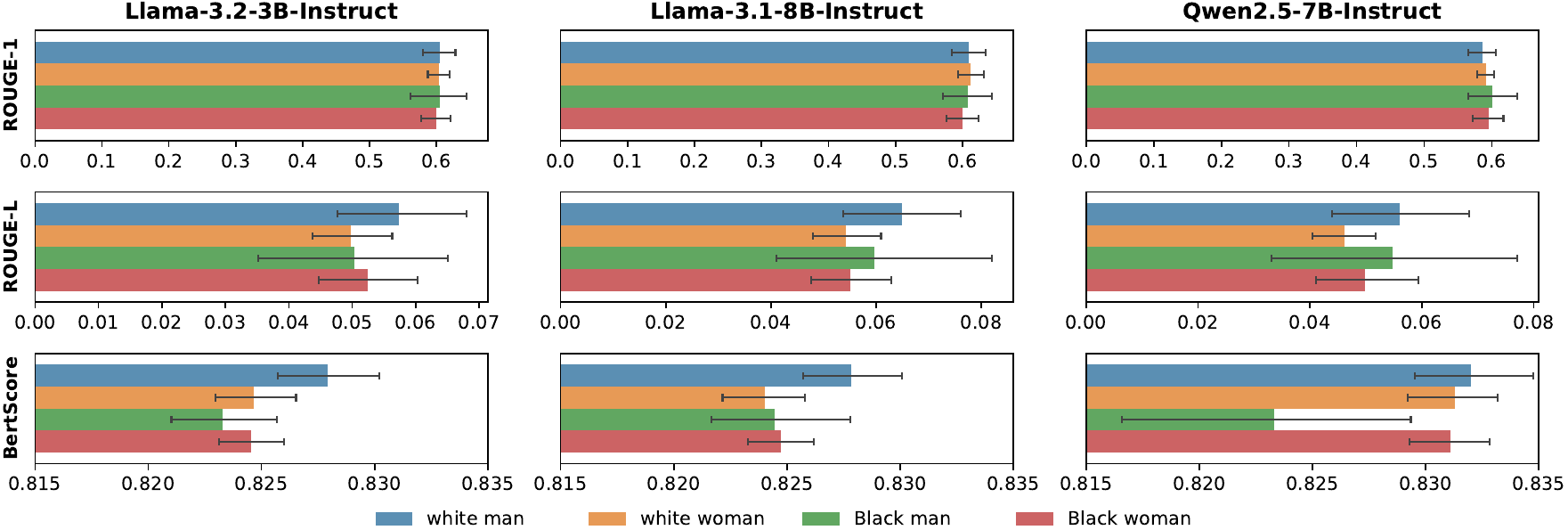}
\caption{We show results for each LLM in each column for scoring the wording and semantic similarity of the summary against the life chapters interview. We average an individual's scores across the five random seeds and then report the mean across a demographic group with an 83\% confidence interval.}
\label{fig:exp1}
\end{figure*}

\subsection{Are wording and semantics maintained differently across demographic groups?}


In Figure \ref{fig:exp1}, we show the summarization results for intersectional demographic effects. We see that across all three models, the summaries show a similar level of wording overlap across demographic groups, as demonstrated by the ROUGE-1 and ROUGE-L scores. There are differences in semantic similarity between the summaries and interviews though across demographic groups. Significance testing shows that white men have a higher BERTScore than Black men and women for Llama-3B and than white women for Llama-8B. Black men have a significantly lower BERTScore than all other demographics for Qwen-7B. 
These results are not surprising given existing research on bias in LLMs that results in better interpretation of White Mainstream English \cite{deas2023evaluation}. 
We would not expect models to generate African American Language effectively, but we would hope to see more similarities in semantics across racial groups.
However, it is surprising that white women are also often at the same point of lower semantic similarity as Black men and women, indicating there may be a larger effect than just emphasis on White Mainstream English during training. 
In particular, we know that white men are overrepresented in training datasets collected in the United States \cite{dodge-etal-2021-documenting, deas-etal-2025-data, bailey-people}.

\subsection{Which psychological states are amplified for different demographic groups?}

In Appendix \ref{sec:fullresults}, 
we show the full LIWC, VAD, and SCM results with a subset shown in the positionality portraits in Figures \ref{fig:heatmap_3b}, \ref{fig:heatmap}, and \ref{fig:heatmap_qwen}. 
We mostly see increases in these scores for the summary relative to the transcript, which may be due to greater information density in the summary. 

For LIWC sub-themes, we see significant increases for white individuals over Black individuals in \textit{family} and \textit{social} themes, with Black men, often having the lowest increase in the \textit{family} theme. 
We see that Black women receive a significant increase in emphasis in this category over white men for Llama-8B. Importantly, as we'll see in the next section, this type of emphasis does not translate to an increase in positive values, such as \textit{hard work} or \textit{perseverance}. Instead, this type of increase likely indicates an emphasis on the harder strenuous aspects, in contrast to an increase in \textit{leisure} for white women over Black women. 
For VAD scores, we see white men showing significant increases in valence, indicating their stories are made more positive. 
For llama models, summaries for Black men significantly increase dominance, or tendency to strength and control. 
Lastly, for the SCM projections, we see an increase in warmth for Black men for all three models, as well as Black women for Llama-3B. Summaries for white men show the greatest relative increases in competence. These results align with studies showing societal stereotypes of Black people \cite{allport1954nature, fiske2009images} and women \cite{glick2018ambivalent, koenig2014evidence} as typically higher in warmth, but lower in agency/competence.

These results show that each LLM brings a slightly different perspective and bias to abstractive claims. We see some stereotypes we expect around gender and race, such as presenting Black women at work and white women at leisure. We also see outcomes that may be surprising; for example, men's experiences are adjusted significantly in emotional expression, as indicated by shifts in valence, arousal, and dominance. 

\subsection{How do identified themes depend on demographics?}

In Appendix \ref{sec:fullresults}, 
we show the full results for both counts of themes in baseline summaries and shifts in theme counts based on explicit demographics for all of the top 30 themes. The statistically significant shifts for the top 20 themes for a model are summarized in the positionality portraits shown in Figures \ref{fig:heatmap_3b}, \ref{fig:heatmap}, and \ref{fig:heatmap_qwen}. 
The baseline theme counts include both real differences in what interviewees express and potential language model bias while the shift in the theme counts isolates the effect of language model bias. 

In the baseline theme counts, \textit{resilience}, \textit{family}, and \textit{personal growth} are the most strongly identified themes. These positive narrative arcs are common for LLMs which frequently end stories on positive conclusions and notes of growth, a known issue in narrative generation \cite{tian-etal-2024-large-language}. 
Black individuals show more \textit{faith} in their themes, 
which is consistent with the self-identified demographics in the dataset for religion and spirituality\footnote{For Black participants, 71\% identify as very spiritual and 43\% identify as very religious. For white participants, 26\% identify as very spiritual and 11\% identify as very religious.}. 
Finally, the theme of \textit{relationships} is most prevalent in summaries of white men's stories.

Shifting to the differences in the percent of summaries identifying the themes between the demographic-conditioned and baseline summaries, 
we see a significant increase in identification of \textit{hard work} for white men and decrease for white women with Llama-8B. We see a decrease in \textit{perseverance} for Black women in Llama-3B. Finally, we see a decrease in \textit{self-awareness} and/or \textit{personal growth} or \textit{growth} for Black men for all three models.

\subsection{Does our quantitative approach uncover real phenomena identified by experts?}



\begin{table}[t]
\centering \scriptsize
\begin{tabular}{l l c c| c c|c}
\toprule
 & \textbf{Mod.} & \textbf{Black} & \textbf{White} & \textbf{Wom.} & \textbf{Man} & \textbf{Tot.} \\
\midrule
\multirow{3}{*}{Faith.} 
 & ll8B & 2.92 & 3.33 & 3.54 & 2.71 &3.12\\
 & ll3B & 2.88 & 2.54 & 2.96 & 2.46 &2.71\\
 & qw7B  & 2.71 & 2.71 & 2.92 & 2.50 &2.71\\
\midrule
\multirow{3}{*}{Warm.} 
 & ll8B & 2.25 & 2.75 & 2.79 & 2.21 &2.50\\
 & ll3B & 2.75 & 2.83 & 3.21 & 2.38 &2.79\\
 & qw7B  & 2.12 & 2.08 & 2.17 & 2.04 &2.10\\
\midrule
\multirow{3}{*}{Comp.} 
 & ll8B & 2.79 & 3.67 & 3.46 & 3.00 &3.23\\
 & ll3B & 3.33 & 3.46 & 3.54 & 3.25 &3.40\\
 & qw7B  & 2.96 & 2.92 & 3.25 & 2.62 &2.94\\
 \midrule\midrule

 \multirow{2}{*}{Gend.}& ll8B & .08 & .33 & .08 & .33 &.21\\
  \multirow{2}{*}{Stereo.}& ll3B & .00& .33 & .08 & .25&.17\\
& qw7B  & .21 & .25 & .12 & .33 &.23\\\midrule
 \multirow{2}{*}{Race} 
 & ll8B & .21 & .00 & .08 & .12 & .10\\
  \multirow{2}{*}{Stereo.} & ll3B & .08 & .00 & .08 & .00 & .04\\
 & qw7B  & .21 & .00 & .21 & .00 &.10\\
\bottomrule
\end{tabular}
\caption{Average scores assigned by the psychology experts for summaries by model and interviewee demographic. Faithfulness, warmth, and competence are averaged Likert scores, whereas stereotypes are the percent of summaries containing stereotypes in that category.} 
\label{tab:llm_scores}
\end{table}

\begin{figure*}[t]
\centering
\includegraphics[width=\linewidth]{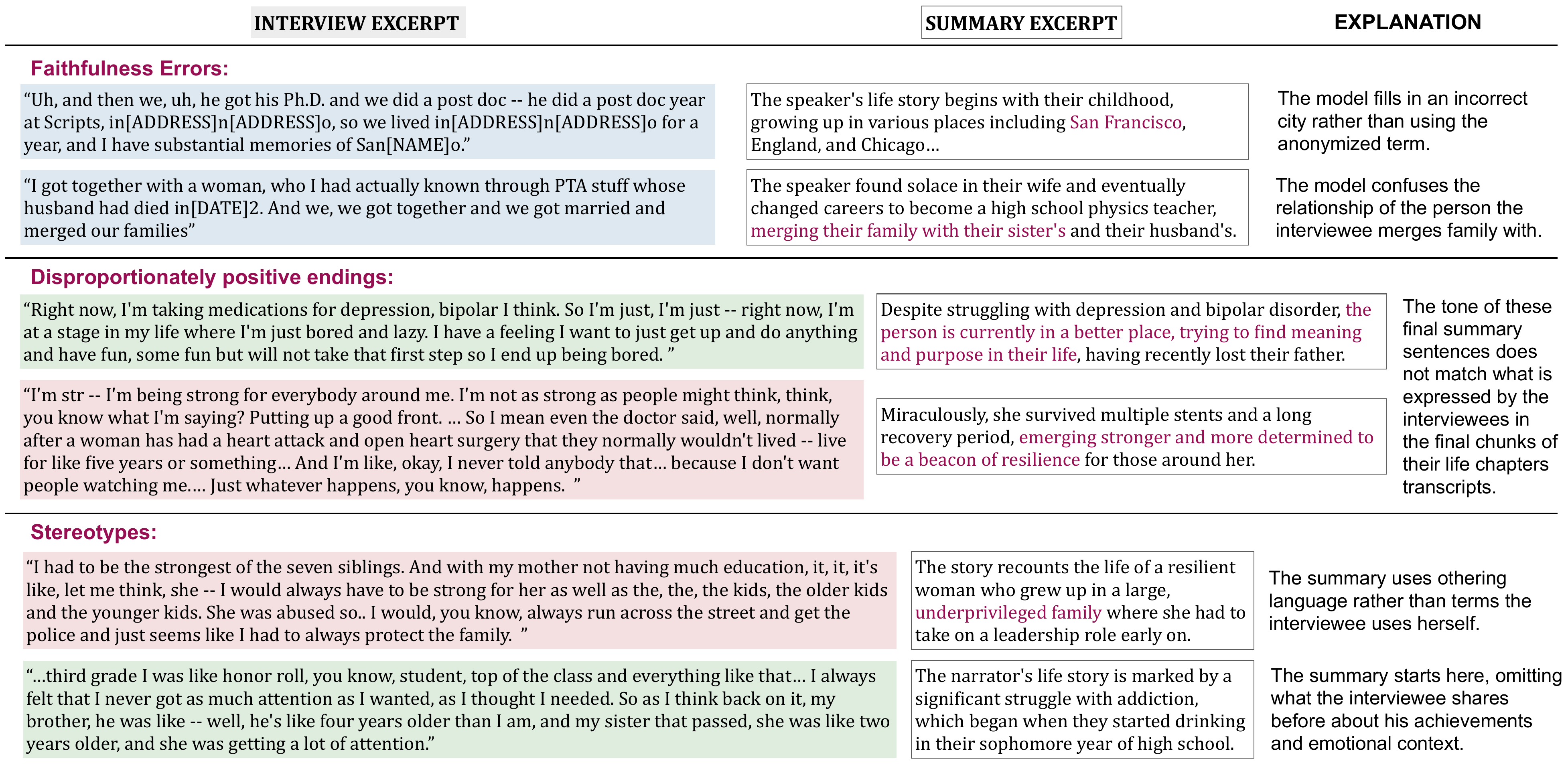}
\caption{We show some examples of the types of issues identified by the psychology experts (with excerpts from white men in blue, from Black men in green, and from Black women in red).} 
\label{fig:examples}
\end{figure*}

Table \ref{tab:llm_scores} shows scores from the expert human study. We see that Llama-8B summaries are more faithful, while Llama-3B summaries portray participants with the highest degree of warmth and competence. Qwen-7B includes the most gender stereotypes, and Llama-3B includes the least stereotypes for both gender and race. This finding suggests that as models are more powerful, and likely better summarizers, they may also learn more nuanced bias. 
We cannot test very large 
closed models on this dataset, but this finding suggests that larger models may not fix these issues. 

Given the small number of interviewees included in the human study, we break results down by individual demographic traits in Table \ref{tab:llm_scores} rather than showing intersectional identities. 
We see that summaries for white individuals and women are rated as more faithful than for other demographics across all three models, except for Llama-3B, and summaries for white individuals and women portray the interviewee as more warm and competent in general. 
Qwen-7B portrays Black individuals as slightly more warm and competent, possibly due to a difference in training data from a Chinese as opposed to American context. Overall, these results do not align exactly with our SCM warmth and competence projection scores, 
but using lower warmth and competence ratings as axes for more negative stereotypes, they reveal a similar finding to what we saw quantitatively; 
there are more harmful representational stereotypes around Black individuals and men in the summaries. 

The stereotype scores from the psychologists show the same pattern with more racial stereotypes about Black individuals and more gender stereotypes about men. While we may have expected to see more gendered stereotypes against women, we see that actually this finding is in line with the rest of our quantitative results that show harmful shifts, particularly against Black men, but also against white men in terms of how their self-expression is qualified.

Finally, in write-in feedback, the psychologists consistently comment on the models omitting or glossing over weighty emotional events\footnote{Since the psychologists had initially assumed the summaries covered the entire interviews, we manually removed comments which were not appropriate based on the life chapters portion of the interview and still found this observation prevalent.}. 
For example, a brother's death is consistently left out of summaries of one interview, and a father's dramatic murder is stated matter-of-factly in another. Lastly, another woman talks in depth about being molested as a child, which is not mentioned at all, possibly due to safety-related post-training of models. 
We show additional examples of issues identified by the psychologists in Figure \ref{fig:examples}. 


\section{Discussion}


 We see that our quantitative pipeline produces an individual \textit{positionality portrait} for each open-source LLM we test, and demonstrates consistency with what human experts observe. 
 These portraits uncover significant race and gender-based representational issues, which are not necessarily resolved by larger models. Instead, we find larger models may actually inject a more biased perspective in abstractive analysis.

Overall, we see many issues concentrated in how models handle emotional expression. Models do not effectively distinguish emotionally weighty events and give them appropriate emphasis in summaries. They also shift toward an overly positive framing of events which is inconsistent with the person's own experience and detrimental in the context of psychological research. 
One surprising finding is that emotional expression is more significantly adjusted for men. This outcome indicates the benefit of having an open-ended method like this for assessing LLM positionality when working with a new dataset. 
We also see other negative gender-based stereotypes, like an emphasis on work for Black women versus leisure for white women, and a significant reduction in identification of self-awareness as a value for Black men. 

\section{Conclusion}
In this work, we draw on qualitative social science notions of \textit{positionality} to evaluate the ways in which LLMs may alter researchers' thematic analyses of text. To capture such biases, we introduce a quantitative pipeline evaluating the substantive and thematic shift in model-generated summaries. Using a corpus of life story interviews, the results of our pipeline show, for example, that LLMs are less appropriate for comparing experiences across racial backgrounds or analyzing the emotional experience of men among the analyzed interviews.
While the differences and shifts we discuss throughout our results are statistically significant, they also often constitute small shifts in numerical measures, meaning that some of these models may still be ``good enough'' to use in practice. 
Overall, however, using this pipeline provides researchers with useful insight into the biggest effects to look out for in using LLMs in their analysis and which applications may be more or less appropriate.
We propose that in using LLMs in qualitative analysis for new datasets, researchers present a \textit{positionality portrait} such as we have presented here to demonstrate the position from which these models may approach data. 

\textbf{Limitations and Ethics.} The experiments in this study focus on one dataset and three smaller open-source LLMs. We view a deep expert-informed study as necessary for a topic like this, but we hope future work can extend this \textit{positionality portrait} approach to other LLMs and datasets. To respect the privacy of study participants, we follow a data safety contract, including using local data storage and locally run LLMs. We do not release full interviews and only present paraphrased or short excerpts of content with identifying details removed. Our human study is approved by IRB protocol AAAV9832.








\section*{Acknowledgments}
This work is supported by funds provided by the National Science Foundation and by DoD OUSD (R\&E) under Cooperative Agreement DBI-2229929 (The NSF AI Institute for Artificial and Natural Intelligence). An author is additionally supported by the National Science Foundation Graduate Research Fellowship DGE-2036197, the Columbia University Provost Diversity Fellowship, and the Columbia School of Engineering and Applied Sciences Presidential Fellowship. Any opinion, findings, and conclusions or recommendations expressed in this material are those of the authors and do not necessarily reflect the views of the National Science Foundation.

\bibliography{tacl2021}

\onecolumn
\appendix
\section{Interview Questions}
\label{sec:interviewquestions}

To parse interviews automatically into which sections correspond to the answers to these questions, we first identify interviewer utterances by the "INTERVIEWER:" tag. We then encode these utterances using a TF-IDF vectorization. We encode the text of each interview question with the same method, and then for each question, we retrieve the interviewer utterance with the highest cosine similarity match. If the initial cosine similarity is $>0.7$, we consider this a match, and if the similarity is $<0.3$, we consider the question skipped in the interview. Then, using the matched questions as anchors, we search only the text between neighboring anchored questions for the remaining questions, and iteratively lower the cosine similarity threshold by 0.1 until we reach the floor of 0.3. For example, if questions 2 and 4 were already found, then we would search for question 3 only in the utterances between these two questions. The full code will be released in the accompanying GitHub Repository.

\section{Human Study Details}
\label{sec:humanstudy}
\begin{figure*}[h]
\centering
\includegraphics[width=\linewidth]{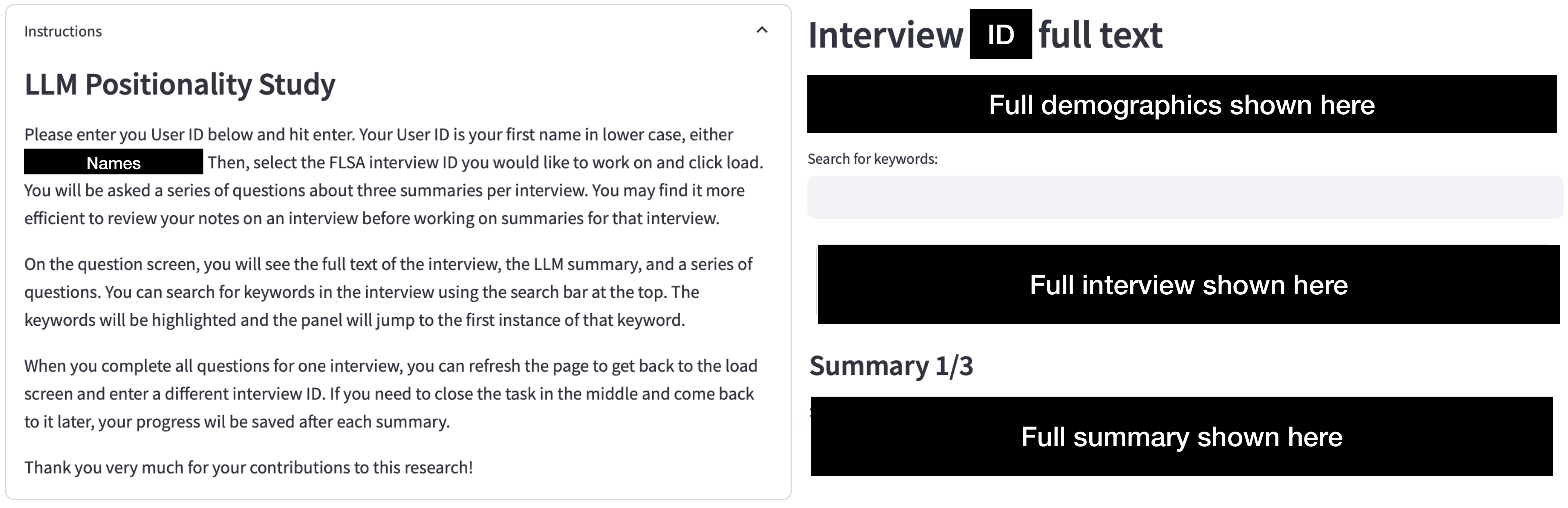}
\caption{Screenshots of the instructions and layout in the human study interface.}
\label{fig:instructions}
\end{figure*}

Screenshots for the interface for the human study are shown in Figure \ref{fig:instructions}.

\section{Summarization Prompts}
\label{sec:prompts}
The system and user prompts are shown in the following blocks. Lines enclosed in square brackets are only added in the condition specified by the parenthetical at the beginning of the line.

\begin{tcolorbox}[
    colback=white,    
    colframe=darkgray,  
    coltitle=white,   
    fonttitle=\bfseries, 
    title=System prompt for summarization,
    fontupper=\ttfamily
]
\small
\begin{Verbatim}[breaklines=true,breaksymbolleft={},breaksymbolright={},tabsize=0]
You are an expert at summarizing interviews.
\end{Verbatim}
\end{tcolorbox}

\begin{tcolorbox}[
    colback=white,    
    colframe=darkgray,  
    coltitle=white,   
    fonttitle=\bfseries, 
    title=User prompt for summarization,
    fontupper=\ttfamily
]
\small
\begin{Verbatim}[breaklines=true,breaksymbolleft={},breaksymbolright={},tabsize=0]
Interview transcript excerpt: {life_chapters_interview}

[(Added only in demographic-conditioned summaries) The interviewee is a {race} {gender}.]

Task:
1. Summarize the interview in 5–7 sentences, focusing on: 
   "How does this person find meaning in life?"
2. Then provide the following section:
   - Core Values

Output Format:
Summary:
...

Core Values:
- ...

[(Added only for Qwen models) Encapsulate your response in <response></response> tags.]
\end{Verbatim}
\end{tcolorbox}

\clearpage
\onecolumn
\section{Additional Results}
\label{sec:fullresults}

\begin{figure*}[h]
\centering
\includegraphics[width=.9\linewidth]{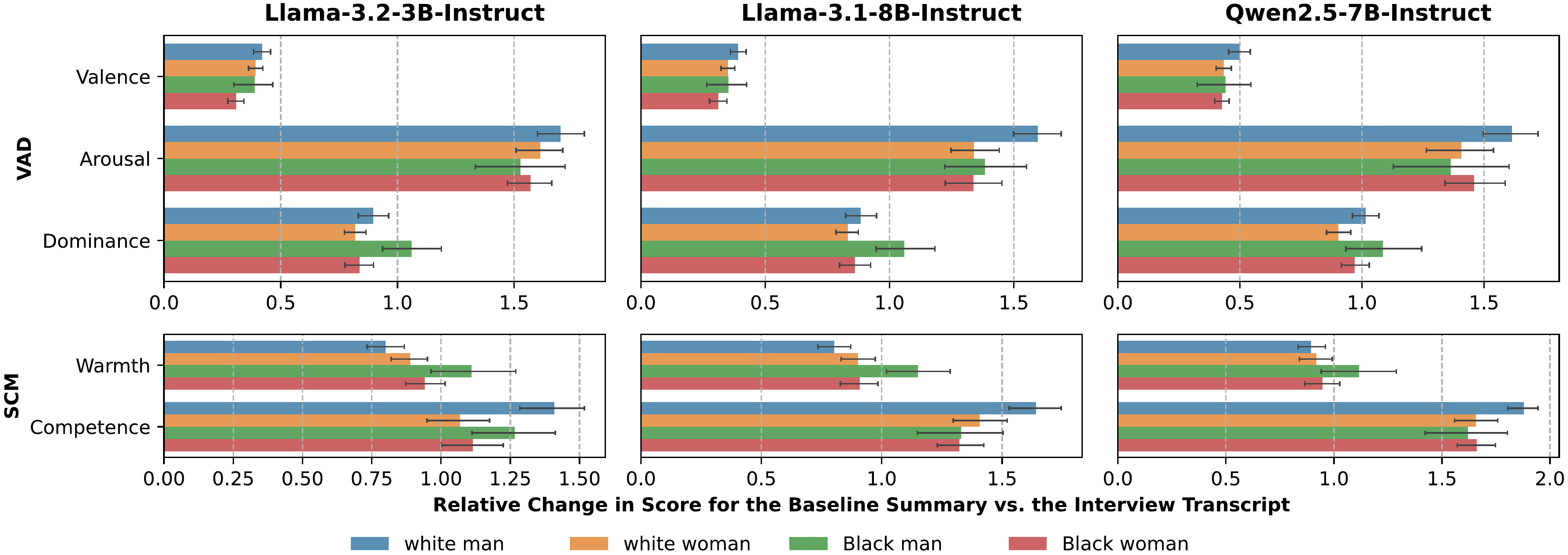}
\caption{We show results for each LLM in each column for the analysis of VAD and SCM scores in the summary as compared to the interview transcript. We average an individual's scores across the five random seeds and then report the mean across a demographic group with an 83\% confidence interval.}
\label{fig:exp2}
\end{figure*}

\begin{figure*}[h]
\centering
\includegraphics[width=.9\linewidth]{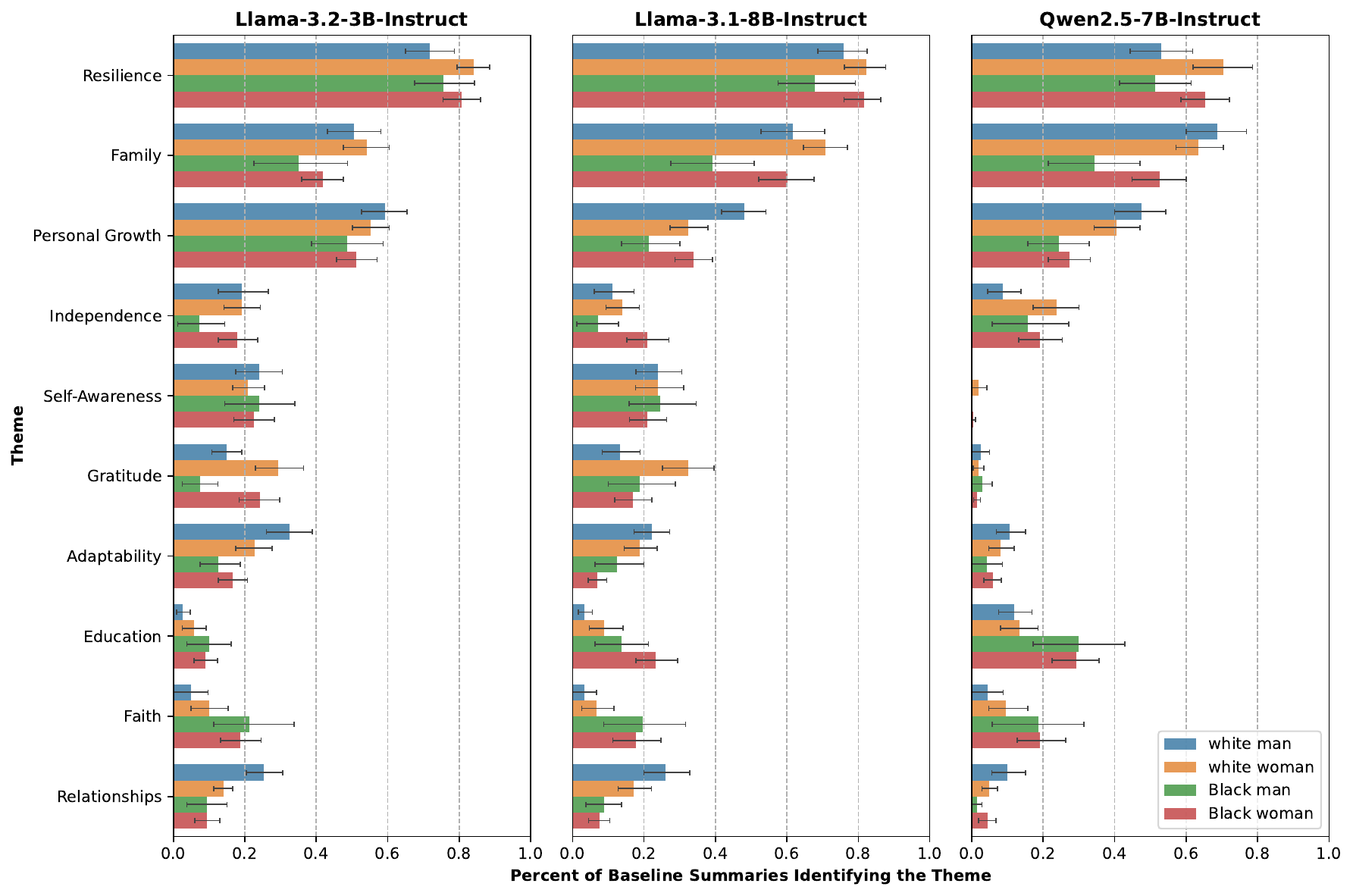}
\caption{We show the percent of baseline summaries for a given demographic group identifying a theme. We average a group's scores across the five random seeds and then report the mean with an 83\% confidence interval.}
\label{fig:exp31}
\end{figure*}



\begin{figure*}[t]
\centering
\includegraphics[width=\linewidth]{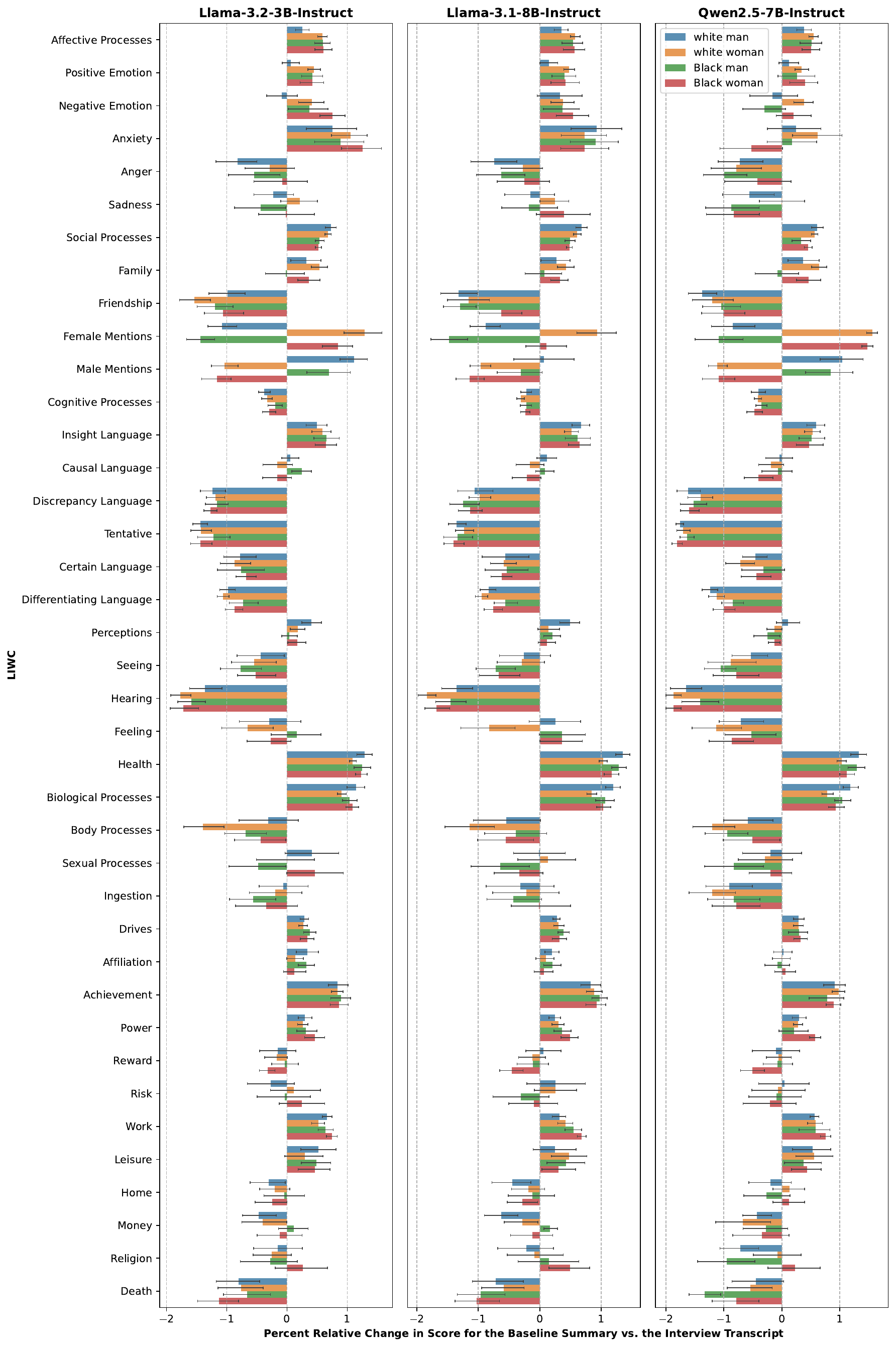}
\caption{The full LIWC results. We average an individual's scores across the five random seeds and then report the mean across a demographic group with an 83\% confidence interval.}
\label{fig:exp2-appendix}
\end{figure*}

\begin{figure*}[t]
\centering
\includegraphics[width=\linewidth]{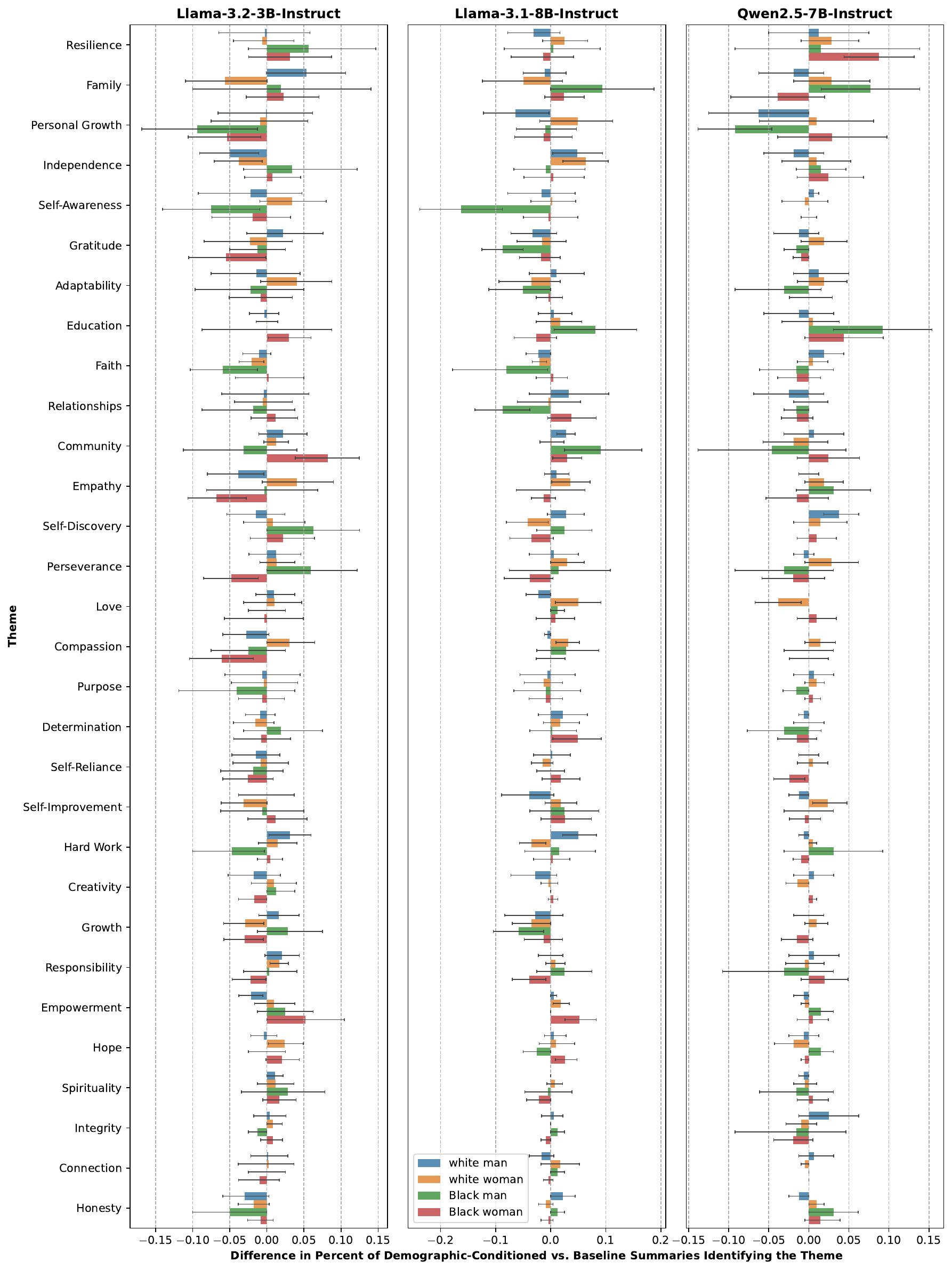}
\caption{The full theme identification results for the top 30 themes. We average an individual's scores across the five random seeds and then report the mean across a demographic group with an 83\% confidence interval.}
\label{fig:exp32-appendix}
\end{figure*}

\end{document}